\documentclass{article}
\usepackage{ijcai25}

\usepackage{times}
\usepackage{soul}
\usepackage{url}
\usepackage[hidelinks]{hyperref}
\usepackage[utf8]{inputenc}
\usepackage[small]{caption}
\usepackage{graphicx}
\usepackage{amsmath}
\usepackage{amsthm}
\usepackage{booktabs}
\usepackage{algorithm}
\usepackage{algorithmic}
\usepackage[switch]{lineno}
\usepackage{enumitem}
\usepackage{array} % For vertical centering in table cells
\usepackage{graphicx} % For \resizebox
\usepackage{multirow}
\usepackage{booktabs}
\usepackage{xcolor}

\title{Multiple Abstraction Level Retrieve Augment Generation}

\author{
Zheng Zheng, Xinyi Ni, Pengyu Hong \\
\affiliations
Department of Computer Science, Brandeis University, Waltham, MA 02453, USA \\
\emails
\{zhengzheng, xinyini, hongpeng\}@brandeis.edu
}
\begin{document}

\maketitle

\begin{abstract}
A Retrieval-Augmented Generation (RAG) model powered by a large language model (LLM) provides a faster and more cost-effective solution for adapting to new data and knowledge. It also delivers more specialized responses compared to pre-trained LLMs. However, most existing approaches rely on retrieving prefix-sized chunks as references to support question-answering (Q/A). This approach is often deployed to address information needs at a single level of abstraction, as it struggles to generate answers across multiple levels of abstraction. In an RAG setting, while LLMs can summarize and answer questions effectively when provided with sufficient details, retrieving excessive information often leads to the 'lost in the middle' problem and exceeds token limitations. We propose a novel RAG approach that uses chunks of multiple abstraction levels (MAL), including multi-sentence-level, paragraph-level, section-level, and document-level. The effectiveness of our approach is demonstrated in an under-explored scientific domain of Glycoscience. Compared to traditional single-level RAG approaches, our approach improves AI evaluated answer correctness of Q/A by 25.739\% on Glyco-related papers. % scientific articles dataset and used GPT-4o-mini to create a 800 Question Answer pairs dataset for evaluation purposes.
\end{abstract}

\begin{figure}[htbp]
    \centering
    \includegraphics[width=0.45\textwidth]{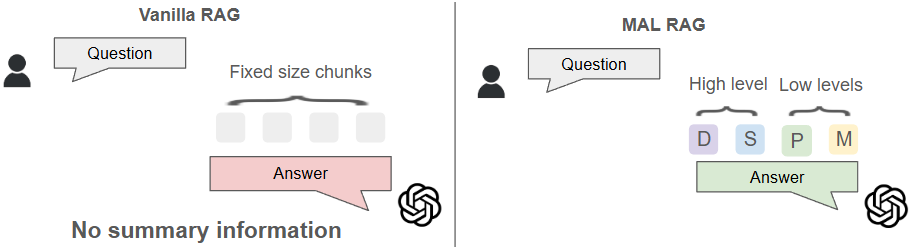}  % Adjust width to fit layout
    \caption{\textbf{Comparison of 
    % \textcolor{red}{Vanilla} 
    Vanilla
    RAG (Left) and MAL-RAG (Right).} In MAL-RAG, $D$, $S$, $P$, and $M$ indicate document-level chunks, section-level chunks, paragraph-level chunks, and multi-sentence-level chunks, respectively. Vanilla RAG, which uses fixed-length chunks, often encounters challenges such as the ``lost in the middle" effect \protect\cite{liu2024lost}. In contrast, MAL-RAG mitigates this problem by utilizing higher-level chunks enriched with summary information.}
    \label{MAL_RAG_pipeline}
\end{figure}

\section{Introduction}

Large Language Models (LLMs)~\cite{brown2020language} have achieved unprecedented success, demonstrating remarkable capabilities in various downstream tasks, which range from traditional NLP applications, such as text classification~\cite{abburi2023generative,zhang2024pushing,zhang2024teleclass}, translation~\cite{koshkin2024transllama,elshin2024general,donthi2024improving}, and summarization~\cite{jin2024comprehensive,ding2024evaluation,pu2023summarization}, to emerging areas, such as code generation~\cite{ugare2024improving,ouyang2023llm} and LLM-assisted decision-making~\cite{eigner2024determinants,chiang2024enhancing}. Despite these advancements, LLMs face significant challenges~\cite{nie2024survey,huang2024comprehensive,yang2024large,ahn2024large,wang2024large} to be solved. The reliance on pretrained LLMs and their black-box nature hinders their ability to generate accurate and specialized responses for professional domains. They may generate incorrect or nonsensical facts (hallucinations~\cite{huang2023survey,feng2024don} and outdated knowledge~\cite{zhang2023large,mousavi2024your}). The professionalism of responses generated by LLMs remains inadequate from the perspective of experts~\cite{ettinger2023you}. RAG is a promising approach to address these challenges~\cite{lewis2020retrieval,guu2020retrieval}. It improves LLMs by combining retrieval and generation, enhancing accuracy with up-to-date, domain-specific knowledge. It scales efficiently by retrieving relevant information at inference and adapts to new data without retraining. RAG provides explainable, evidence-based responses and supports domain expertise through specialized datasets. It reduces costs by minimizing retraining, ensures contextual relevance, and mitigates hallucinations. Customizable and capable of handling multi-modal data, RAG is well-suited for applications requiring accurate, adaptable, and context-aware responses.

However, existing RAG approaches tend to utilize external knowledge from a single perspective, referring to fixed-size chunks in generating answers (e.g., LangChain\footnote{LangChain:\url{https://python.langchain.com}} and LLamaIndex\footnote{LLamaIndex: \url{https://www.llamaindex.ai/}}), which can lead to the extraction of fragmented and/or incomplete information. In real applications, the information needs of a user can be of various abstraction levels, which cannot be fulfilled by individual chunks. For example, to correctly answer a user question, one may need to refer to a whole section rather than a paragraph in a document. Although advanced RAG methods have improved chunking techniques, they have yet to effectively utilize the inherent abstraction structures of reference documents. One may retrieve multiple chunks to address this problem. However, determining the appropriate number of chunks to retrieve remains an open question. Retrieving an excessive number of chunks can degrade the performance of LLMs by introducing excessive noise, which could mislead LLMs~\cite{jin2024long} and cause the “lost in the middle” issue~\cite{liu2024lost}.

% It remains an open question how many chunks should be retrieved. Retrieving excess chunks can severely degrade the performance of large language models (LLMs) due to the rapidly increased noise within the retrieved content, misleading LLMs \cite{jin2024long} and leads to a “lost in the middle” issue \cite{liu2024lost}, as the retrieved documents are too long.

Our contributions: To address the aforementioned challenges, we propose the Multiple Abstraction Level Retrieval-Augmented Generation (MAL-RAG) framework, designed to enhance question reasoning in scientific domains. Following the framework, we implemented a pipeline for reading, parsing, indexing, and segmenting domain-specific literature by leveraging the inherent structures of scientific papers, enabling the construction of a high-quality database that indexes scientific papers hierarchically at multiple abstraction levels. By leveraging the MAL indexing of original content, MAL-RAG enhances the comprehension of complex scientific articles. We applied MAL-RAG to a meticulously curated dataset of Glyco-related papers and demonstrated its superiority over the standard RAG approach.

% To solve the challenges mentioned above, we propose the Multiple Abstraction Level Retrieval-Augmented Generation (MAL-RAG) framework. , integrated with Large Language Models for biological and chemical question reasoning, specifically related to Glyco topic. With a comprehensive pipeline for reading, parsing, indexing, and segmenting the Glyco-related papers, the MAL-RAG framework constructs a high-quality hierarchical relation nodes database, which can be easily extended to other scientific domains. MAL-RAG improves the understanding of complex articles by combining the inherent structural relationships in the content with prebuilt multi-level abstraction nodes. We illustrate the performance of MAL-RAG on manually collected Glyco-related papers and compare it with a native RAG method, relying on a GPT-4o-mini generated QA dataset with suitable prompt simulating the researchers' query process.

%The framework selects relevant chunks using the open source embedding model Linq-Embed-Mistral \cite{LinqAIResearch2024}. 

\section{Related Works}

\subsection{RAG}
Retrieval-Augmented Generation (RAG), introduced by Lewis et al.~\cite{lewis2020retrieval}, promoted the application of LLMs across multiple tasks by providing external information to the model~\cite{borgeaud2022improving,li2024empowering,khandelwal2019generalization,min2020ambigqa,izacard2020leveraging}. Modern LLM frameworks, such as LangChain and LlamaIndex, offer foundational RAG implementations. These typically involve converting documents into text chunks, embedding them as indices, and retrieving semantically relevant references during inference. Beyond these basic implementations, advanced RAG methodologies have been developed to address specific challenges and optimize different stages of the retrieval process. These improvements can be categorized into pre-retrieval and post-retrieval enhancements. In the pre-retrieval process, previous works are focusing on optimizing the indexing structure and the original query, using technologies such as data granularity, optimizing index structures, adding metadata~\cite{hayashi2024metadata,ghasemi2024harnessing}, a query rewriting~\cite{mao2024rafe,li2024dmqr,ma2023query,peng2024large,ma2023query}, query transformation~\cite{chan2024rq}, and query expansion~\cite{zheng2023take,gao2022precise}. For post-trieval process, the critical problem to solve is determining whether the retrieved context is effective for the query~\cite{jin2024llm}. These techniques include re-ranking the retrieved information~\cite{hwang2024dslr,mishra2024searchd,glass2022re2g}, emphasizing critical sections~\cite{shi2024enhancing,csakarmaximizing,shi2024enhancing}, and shortening the context to be processed~\cite{zhang2024long,bai2024longalign,zhao2024longrag}.

\subsection{Scientific Domain-Specific RAG}
As RAG LLM techniques are beneficial for domain-specific scientific areas, they have been widely adopted in fields such as medicine, biology, and finance. In the medical domain, Lozano et al.~\cite{lozano2023clinfo} proposes an open-source RAG-based LLM system designed for answering medical questions using scientific literature. Similarly, Wang et al.~\cite{wang2024biorag} introduce a robust pipeline comprising document parsing, indexing, segmentation of extensive research papers, embedding model training, and LLM fine-tuning to enhance performance. Jiang et al.~\cite{jiang2024tc} integrate a Turing Complete System for efficient document retrieval and management, enabling accurate responses to medical queries. Additionally, some RAG systems have been extended to molecular research by integrating the retrieval of molecular structures and biomedical entities such as proteins, molecules, and diseases~\cite{liu2023multi,wang2022retrieval,wang2023biobridge,yang2023prompt}. In the financial sector, Lin~\cite{lin2024revolutionizing} proposed a PDF parser combined with RAG-based LLMs to extract knowledge from financial reports. Yepes et al.~\cite{yepes2024financial} introduced a novel document chunking approach based on structural elements rather than traditional paragraph-based chunking, enhancing the retrieval process. Despite their widespread application, these RAG systems rely on basic methods with fixed-size chunks and lack attention to the completeness and higher-level background information of the retrieved contexts.

\begin{figure*}[htbp]
    \centering
    \includegraphics[width=\textwidth]{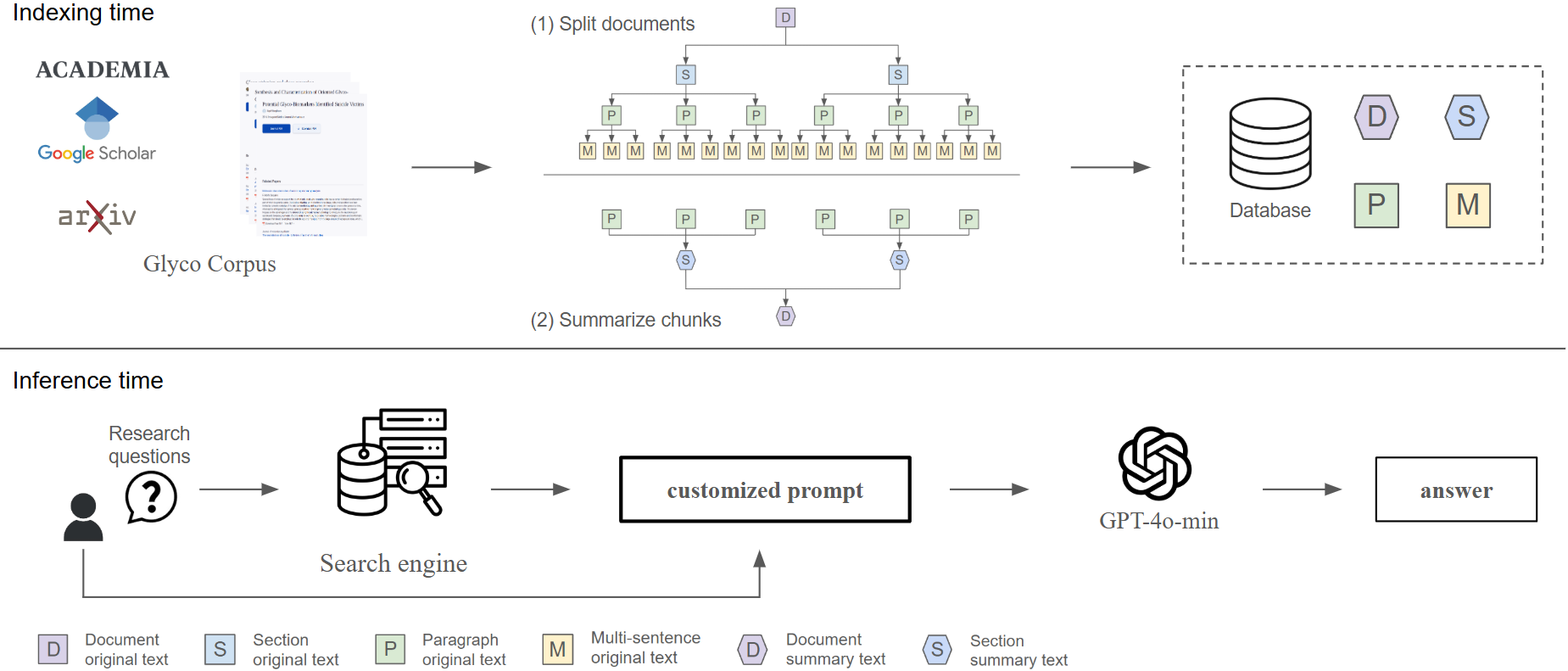}
    \caption{\textbf{MAL-RAG Pipeline.} The MAL-RAG pipeline is composed of two primary stages: indexing and inference. In the indexing stage, articles are divided into multiple levels of granularity, such as document-level, section-level, paragraph-level, and multi-sentence-level text. A map-reduce approach is then used to extract key information from paragraph-level chunks, which are summarized into section-level chunks. These section-level chunks are further processed to generate document-level chunks in a similar manner. In the inference stage, a search engine retrieves relevant chunks based on similarity scores, which are computed using the Linq-Embed-Mistral open-source embedding model. These retrieved chunks, along with the input question and prompts, are fed into GPT-4o-mini to generate the final response.}
    \label{MAL_RAG_pipeline}
\end{figure*}

\subsection{Chunking Optimization}
Focusing on the quality of retrieved chunks and effectively capturing background information, many chunking strategies have been proposed for optimization. Common approaches include fixed-size chunking, recursive chunking, sliding window chunking, paragraph-based chunking, and semantic chunking. While longer text chunks preserve more semantic coherence, they can also introduce noise, dilute the model's attention, and lead to the "lost in the middle" phenomenon~\cite{liu2024lost}. Advanced methods address these challenges by dynamically determining the appropriate level of detail and selecting chunks with optimal granularity~\cite{sarthi2024raptor,zhong2024mix,chen2024hiqa}. Other approaches refine text into smaller, information-rich segments to maintain high completeness~\cite{wu2021recursively,angelidis2018summarizing,chen2023dense}. For instance, Gao et al.~\cite{gao2023enabling} showed that summarizing passages into shorter text improves accuracy, while Zhao et al.~\cite{zhao2024longrag} proposed LongRAG, which condenses retrieved contexts into summaries that balance informativeness and conciseness. Similarly, Edge et al.~\cite{edge2024local} used LLMs to construct graph representations of corpora, creating detailed nodes and summaries for retrieval. Although these strategies achieve state-of-the-art performance, many rely on model predictions or clustering to group chunks, generating higher-level abstractions or summaries. In practice, chunks sequentially sourced from the same section or document, especially when following the intrinsic structure of scientific documents, often align better with human summarization tendencies than those produced by models. To address this issue, our MAL-RAG leverages the intrinsic structure of documents to group chunks in a manner that more naturally aligns with human reasoning, constructing higher-level abstractions that improve both coherence and retrieval accuracy.

\section{Multiple Abstraction Level Retrieval-Augmented Generation Framework}
%This section explains the technical details of our MAL-RAG framework.
%\subsection{Problem Formulation}
Our MAL-RAG framework aims to utilize the native structures of articles to answer questions that require information at different levels of abstraction. When building indexes, we preprocess each reference document $d \in \mathcal{D}$ into four level chunks: document-level $\mathcal{D}_c$, section-level $\mathcal{S}_c$, paragraph-level $\mathcal{P}_c$, and multi-sentence-level $\mathcal{M}_c$. During the inference time, we apply a retriever $\mathcal{R}$ to utilize different abstraction-level chunks. To balance information richness and retrieval efficiency,  the method dynamically adjusts the number of chunks extracted until the length of the accumulated text reaches a predefined range $r$. Subsequently, the probability of each chunk is calculated using a similarity-based softmax approach. Chunks are iteratively accepted until the accumulated probability reaches a pre-specified threshold $p$. After the relevant chunks are retrieved, the question and chunks are fed into an Answer Generator to generate the answer. The technical details are explained in the following subsections.

\subsection{MAL-RAG Chunks Generation}

As illustrated in Figure \ref{MAL_RAG_pipeline}, chunks are designed at four levels of abstraction, corresponding to varying granularities of the original documents, which are document-level chunks $\mathcal{D}_c$, section-level chunks $\mathcal{S}_c$, paragraph-level chunks $\mathcal{P}_c$, and multi-sentence-level chunks $\mathcal{M}_c$. During the generation process, the articles are split into these four levels based on their inherent structure. 

For document-level and section-level chunks, directly using the original content extracted from the document can result in excessive length and dilute the LLM's attention. To address this, we employ a map-reduce approach to generate summarized information. In this approach, we first produce summaries for each paragraph. These paragraph summaries are then used to generate the section-level summary, which serves as the content for the section-level chunks. Finally, we aggregate these section summaries to create document summaries, which serve as the content for the document-level chunks.

For all summary tasks in the generation of document-level and section-level chunks, we instantiated $\mathcal{E}$ as the pre-trained large language model Vicuna-13B-v1.3~\cite{zheng2023judging} with customized prompts designed to summarize the key information from the provided contexts. Discussion of the prompt details is included in supplementary material section A. 

For paragraph-level and multi-sentence-level chunks, we retain the original content and store it directly in the chunk database to provide detailed information during inference.

Given a document $d_i \in \mathcal{D}$, which contains $N_i$ sections $s_{i,1}, s_{i,2}, \cdots, s_{i,N_i}$. Consider an arbitrary section $s_{i,n} \in d$ that consists of $K$ paragraphs $p_{i,n,1}, p_{i,n,2}, \cdots p_{i,n,K}$. The key information of $s_{i,n}$, denoted as $\tau_{i,n}$ is defined as:

\begin{equation} 
    \tau_i = \mathcal{E}\left(\sum_{j=1}^{k} \mathcal{E}(p_{i,j})\right)
\end{equation}
where $\mathcal{E}(\cdot)$ is the key information extractor applied to each paragraph in $s_i$ and the subsequent aggregation over all paragraphs in the section.

Similarly, the key information of the document $d$ is defined as:

\begin{equation}
    d = \mathcal{E}\left(\sum_{i=1}^{n} s_{c_i}\right), \quad i \in {1,\cdots, n},
\end{equation}
where the aggregation is performed over the key information of all $n$ sections in the document. Finally, we construct the chunk database, which contains all chunks from four abstraction levels.

\subsection{Chunk Retrieval}

Given a question and a chunk, the retriever applies the embedding model Linq-Embed-Mistral~\cite{LinqAIResearch2024} to generate the embedding vector $q$ for the question and $c$ for the chunk. The similarity between the question $q$ and the chunk $c$ is computed by the cosine similarity:
\begin{equation}
    \text{Sim}(q, c) = \frac{q \cdot c}{||q|| ||{c}||}, \quad c \in D_c \cup S_c \cup P_c \cup M_c
\end{equation}

This process is repeated until the similarity scores are calculated with all chunks. The retriever selects the most relevant chunks while adhering to a length constraint $C$, which specifies the maximum total length of the selected chunks. To further optimize the retrieval process and minimize noise, we employ a softmax equation to convert similarity scores into probabilities.  Suppose \( k \) chunks are selected. For an arbitrary chunk \( c_i \), the probability is defined as:

\[
P(c_i \mid q) = \frac{\exp(\text{Sim}(q, c_i))}{\sum_{j=1}^{k} \exp(\text{Sim}(q, c_j))},
\]
The chunks are sorted in descending order based on their probabilities. Top chunks, whose cumulative probability does not exceed a pre-defined threshold $\tau$, are selected for generating an answer.

%\begin{equation}
%    \underset{S}{\operatorname{argmax}} \,\sum_{i \in S} \text{length}(c_i) \leq C,
%\end{equation}

%where $S$ is the set of maximum number of selected chunks within the limited length $C$. 

%\begin{equation}
%    \underset{S}{\operatorname{argmax}} \sum_{i \in S} P(c_i \mid q) \leq \tau,
%\end{equation}
%where $S$ is the set of maximum number of selected chunks within the limited probability $\tau$. This ensures that the %retriever balances relevance and coverage while adhering to the length constraint.

\subsection{Answer Generation}
The retrieved chunks are concatenated and represented as $C_s$. We carefully prompt 
% \textcolor{red}{[Which LLM:] updated the details about the LLM}
with the open-source LLM, Vicuna-13B-v1.3~\cite{zheng2023judging}, to generate an answer:
\begin{equation}
a = \text{LLM}(\text{prompt}(C_{s}, q)),
\end{equation}
where $\text{prompt}(C_s, q)$ encapsulates the context (i.e., $C_s$) and the question ($q$), ensuring the LLM can produce a coherent and accurate response. In this prompt, we applied the ICL method to further guide the LLM in generating the desired data. The details about this prompt can be found in Section B of the supplementary material.
% \textcolor{red}{[Explain the prompt design:]}

\subsection{Metrics for Evaluating Answers}

We assess the quality of the answers generated by the LLM using a set of metrics implemented in the Ragas package\footnote{\url{https://docs.ragas.io/en/v0.1.21/concepts/metrics/index.html}}, including Faithfulness, Answer Relevancy, Answer Similarity, Answer Correctness, Context Precision, Context Utilization, Context Recall, and Context Entity Recall. When calculating the metrics, we split both the ground truth and the generated answers into sentences. Each sentence is treated as a statement, and an LLM model is introduced to assess whether two statements match. In this research, we used GPT-4o-mini for this task.
% \textcolor{red}{[How to decide two statements match each other? Updated]}
The primary evaluation metric is Answer Correctness, which is measured by the F1 score:

\begin{equation}
    F1\text{ score} = \frac{|TP|}{|TP| + 0.5 \times (|FP| + |FN|)}
\end{equation}
where TP refers to the statements that are present in both the ground truth and the generated answer, FP represents the statements that are present in the generated answer but not in the ground truth, FN denotes the statements that are present in the ground truth but missing in the generated answer. 

\section{Experiment Results}
\subsection{Experimental Setup}

\paragraph{Dataset} 
We constructed a set of 7,652 academic articles in English that are closely relevant to Glycoscience or Glycomaterials. After preprocessing, the chunk database contains 7,652 document-level chunks, 138,259 section-level chunks, 494,613 paragraph-level chunks, and 1,176,259 multi-sentence-level chunks (see Table \ref{table:stats}). 

\begin{table}[t]
\centering
\begin{tabular}{lcc}
\toprule
\textbf{Chunks Level}          & \textbf{Num of $chunk$} & \textbf{Avg. Length of $chunk$} \\ 
\midrule
\textbf{Document}       & 7652 & 347 \\
\textbf{Section}       & 138259 & 375 \\
\textbf{Paragraph}       & 494613 & 876 \\
\textbf{Multi-sentence}       & 1176259 & 338 \\ 
\bottomrule
\end{tabular}

\vspace{0.25cm}

\caption{Statistics of chunks for different levels. "Avg. Length" refers to the average word count.}
\label{table:stats}
\end{table}

\begin{table*}[htbp]
\centering
\setlength{\tabcolsep}{3mm} % Adjust column spacing for better fit
\renewcommand\arraystretch{1.5} % Adjust row height for better readability
\resizebox{\textwidth}{!}{
\begin{tabular}{cccccccccc}
\toprule
\multicolumn{2}{c}{\textbf{Approaches}} & \textbf{Faithfulness} & \textbf{Answer Relevancy} & \textbf{Answer Similarity} & \textbf{Answer Correctness} & \textbf{Context Precision} & \textbf{Context Utilization} & \textbf{Context Recall} & \textbf{Context Entity Recall} \\ \midrule
\multicolumn{2}{c}{\textbf{RAG w. c}} & 86.317 & 90.15 & 95.21 & 82.04 & 94.978 & 95.426 & 92.481 & 68.379 \\ \midrule
\multicolumn{2}{c}{\textbf{Vanilla RAG}} & 71.982 & 83.557 & 88.707 & 43.049 & 59.246 & 80.252 & 27.734 & 9.721 \\ \midrule
\multirow{2}{*}{\textbf{Document RAG}} & $\tau=0.5$ & 47.401 & 59.942 & 84.998 & 40.925 & 43.853 & 53.779 & 23.435 & 14.353 \\ 
& without $\tau$ & 50.934 & 85.487 & 41.773 & 41.360 & 48.648 & 26.755 & 15.544 & 15.377 \\ \midrule
\multirow{2}{*}{\textbf{Section RAG}} & $\tau=0.5$ & 73.313 & 84.234 & 91.183 & 57.245 & 71.512 & 80.367 & 55.128 & 33.936 \\
& without $\tau$ & 77.115 & 87.776 & 91.540 & 57.489 & 65.302 & 74.785 & 63.602 & 35.226 \\ \midrule
\multirow{2}{*}{\textbf{Paragraph RAG}} & $\tau=0.5$ & 80.497 & 88.023 & 91.838 & 59.090 & 67.191 & 78.804 & 64.934 & 32.758 \\
& without $\tau$ & 85.653 & 91.780 & 92.374 & 59.450 & 62.797 & 72.184 & 73.491 & 33.577 \\ \midrule
\multirow{2}{*}{\textbf{Multi RAG}} & $\tau=0.5$ & 81.366 & 90.339 & 92.451 & 63.046 & \textbf{82.414} & 63.946 & 35.907 & 36.494 \\
& without $\tau$ & 86.226 & 92.911 & 92.729 & 61.674 & 64.498 & 75.103 & 73.751 & 37.301 \\ \midrule
\multirow{2}{*}{\textbf{MAL-RAG}} & $\tau=0.5$ & 86.246 & 91.235 & 93.438 & \textbf{68.788} & 80.553 & \textbf{87.552} & 79.100 & \textbf{51.560} \\
& without $\tau$ & \textbf{89.228} & \textbf{93.802} & \textbf{93.630} & 66.661 & 75.483 & 81.794 & \textbf{86.061} & 51.232 \\ \bottomrule
\end{tabular}
}
\vspace{0.25cm}
\caption{Performance evaluation of different RAG strategies. The table presents the scores across various metrics, including faithfulness, answer relevancy, similarity, correctness, and context-related factors (precision, utilization, recall, and entity recall). These metrics are computed using the Ragas framework.}
\label{tab:result}.
\end{table*}

\paragraph{Evaluation Questions/Answer Dataset}
To evaluate the effectiveness of the RAG system for a customized database lacking human-curated Q/A datasets, we generated a dataset of 1,118 Q/A pairs using GPT-4o-mini and selected 200 pairs from each level, totaling 800 pairs for the evaluation dataset. Each question in this dataset is a short phrase-answer query, where the task of the large language model (LLM) is to provide an answer based on the given question and the retrieved context. These Q/A pairs were derived from randomly selected articles, with 200 Q/A pairs assigned to each granularity level, employing distinct generation strategies for each level. At the document-level, 
% \textcolor{red}{[If 3 questions are generated for each chosen document chunk, how to get 200 Q/A at the document level?: Updated above]} 
three questions were generated for each selected document chunk.
% \textcolor{red}{[Don't understand this part. What are those 7 most common section titles? Usually, the titles are very short. How to use them to generate meaningful Q/A pairs?: Updated with explanation]}

At the section-level, we identified seven types of sections, including the titles 'Introduction', 'Discussion', 'Conclusions', 'Conclusion', 'Statistical Analysis', 'Results and Discussion', and 'Results'. These sections provide comprehensive and crucial information, are commonly found in research papers, and can be easily used to generate higher-level abstract questions that are more representative of the content in scientific literature.
For the paragraph-level and multi-sentence-level, since different chunks contain varying levels of specialized knowledge in glycoscience, we aim to ensure that our questions more accurately reflect those posed by experts in the glyco field, rather than general questions that might arise from other domains. To achieve this, we employed a fine-tuned model based on GIST-small-Embedding-v0~\cite{solatorio2024gistembed} to predict the relevance of content to the Glyco-related corpus.
% \textcolor{red}{[Why is this necessary? Aren't all chunks from Glyco-related papers?: Updated with explanation]}
The model was trained using the SciRepEval dataset~\cite{singh2022scirepeval} and our own Glyco-related abstracts. Specifically, we utilized the Fields of Study (FoS) task\footnote{The\thinspace Fields\thinspace of\thinspace Study\thinspace task\thinspace in\thinspace the\thinspace SciRepEval\thinspace dataset:\\\url{https://huggingface.co/datasets/allenai/scirepeval/viewer/fos}} from the SciRepEval dataset, which is designed for scientific document representation. To adapt the model for Glyco-related content, we re-labeled the dataset used. Labels from our collected abstracts were assigned a value of 1 to indicate relevance to the Glyco domain, while labels from the FoS dataset abstracts were assigned a value of 0. This transformed the task into a binary sequence classification problem. We trained the model using the first 7,000 tokens of the abstracts. The model, GIST-small-Embedding-v0, was fine-tuned, and the softmax layer output was then used as the final probability for each paper's relevance to the Glyco domain.
% \textcolor{red}{[how were they used: Update that I used gpt-4o-mini to do this]}
Based on these relevance scores, we selected the top 5 chunks with the highest scores from each granularity level, considering them the most representative examples. These selected chunks were then used to generate QA pairs with GPT-4o-mini.

% \paragraph{Metrics}
% \textcolor{red}{[What is the purpose of this paragraph?] Recent advances in large language models (LLMs) have demonstrated state-of-the-art or competitive performance in evaluating natural language generation tasks, often surpassing or matching human judgment~\cite{wang2023chatgpt}. For instance, LLMs like GPT-4 achieve over 80\% agreement with human evaluators, reaching a level of alignment comparable to human assessments~\cite{zheng2023judging}. Additionally, LLMs have shown considerable promise in evaluating conventional Retrieval-Augmented Generation (RAG) systems, particularly in metrics such as faithfulness, answer relevance, and reference recall.}

% \textcolor{red}{[This is redundant to Section 3.4] In our evaluation, we employed the Ragas package, which includes a comprehensive set of metrics for assessing the quality of answers and retrieved contexts. These metrics include Faithfulness, Answer Relevancy, Answer Similarity, Answer Correctness, Context Precision, Context Utilization, Context Recall, and Context Entity Recall. This holistic approach allows for a thorough evaluation of both the generated responses and the relevant context used during retrieval.}

\paragraph{RAG approaches} We compare the performance of MAL-RAG with several other RAG approaches, each utilizing the GPT-4o-mini model to generate answers: 

\begin{enumerate}[label=\roman*.]
\item \textbf{Vanilla RAG}: This approach implements a basic Retrieval-Augmented Generation (RAG) setup using the sentence-splitter retriever and generator from the LLama-Index package. Unlike sentence-level chunking methods, in which documents are divided into smaller semantic units such as sections and paragraphs, the sentence-splitter in this configuration solely focuses on splitting the document into individual sentences. These sentences are then merged iteratively until they approximate a pre-defined chunk size. Different from multi-sentence-level RAG, this method does not consider the semantic coherence between sentences, which can result in the grouping of sentences from different paragraphs or sections and can lead to incoherent or contextually fragmented chunks. 
% \textcolor{red}{[Does it mean it works at sentence-level?:explained in details about the difference between Vanilla RAG and sentence-level RAG]}

\item \textbf{RAG with Corresponding Chunks}: 
% \textcolor{red}{[Don't understand this. Do you mean this approach does not need a chunk retriver as the ground truth chunk is used?: Update with explanation of what the chunks are used and why]} 
This method uses ground truth as the retrieved chunks to evaluate the performance of GPT-4o-mini when provided with correctly retrieved information for the questions. This serves as a benchmark to compare the effectiveness of other retrieval strategies.

\item \textbf{Single-Abstraction-Level RAG}: 
% \textcolor{red}{[How to decide which level should be used?: As all levels are used individually, there is no need to decide which level should be used]} 
We sequentially use chunks from a single level as the retrieval source to evaluate the effectiveness of chunks at different levels in the questions. These levels are the document-level, section-level, paragraph-level, and multi-sentence-level, which are shown in Table \ref{tab:result}.
% Explain what are single-abstraction levels.

% \item \textbf{MAL-RAG}: This is our approach retrieves and utilizes all chunk levels to assess the performance of the MAL-RAG system.
\end{enumerate}

To ensure comparable information is used by different RAG approaches to answer questions, we set the retrieval context length $C$ to a maximum of 10,000 words. In addition, we explored two settings in retrieving chunks: one sets the probability threshold $\tau = 0.5$, and the other does not have this constraint.

\subsection{Performance}

The experimental results (see Table \ref{tab:result}) demonstrate the effectiveness of our MAL-RAG approach on the Question-Answering dataset derived from scientific papers in the Glyco-domain. The comparison results indicate that the MAL-RAG strategy outperforms single-perspective approaches in multiple metrics, showing either the best performance or competing closely with the best results. When compared to vanilla RAG, it is evident that MAL provides more accurate and complementary information that is relevant to the required answers.

Furthermore, in the ablation experiments, to more clearly analyze the advantages of multiple levels in the retrieval process, we isolated individual levels for retrieval. The comparison shows that as the content becomes more detailed (from document-level to multi-sentence-level), the evaluation scores improve. However, MAL-RAG still shows a significant performance boost of 1\% to 7\% compared to other single-level retrieval methods. This demonstrates that multiple-dimensional perspectives provide information that other levels cannot, making MAL-RAG more effective than other strategies.

Additionally, considering that MAL-RAG is more likely to be affected by noise when using a larger number of chunks compared to other strategies, we employed similarity measures to assess the effectiveness of chunks in relation to the query. Softmax normalization was applied to mitigate the impact of different chunks. To further reduce noise in the retrieval process that could affect the quality of the LLM's answers, we introduced a threshold for accumulating probability, $p$. By comparing the performance with and without a threshold (set at 0.5), we observed that this strategy improved answer correctness by approximately 2\% and enhanced relevance. However, it also led to a slight decrease in context recall.

\section{Limitation}
% \textcolor{red}{[I don't understand the logic:] }
% The experimental results reveal that multi-sentence chunks deliver competitive performance. This suggests that, given sufficient and relevant details, LLMs inherently summarize content as part of the question-answering process. However, by explicitly incorporating summarization, our MAL-RAG framework enables LLMs to access higher-dimensional information using fewer tokens, thereby mitigating the challenges associated with long-context scenarios. Consequently, higher-dimensional chunks can serve as an effective strategy for improving performance when the available tokens are limited.
In this research, the document-level summaries and section-level summaries were derived by the open source LLM model Vicuna-13B-v1.3~\cite{zheng2023judging}. Hence, their quality entirely depends on the summarization capability of Vicuna-13B-v1.3. Document-level summaries, in particular, are more susceptible to the model's limitations as they are generated by aggregating the summaries of sections within each document. % Furthermore, the quality of document-level summaries is inherently reliant on the accuracy and coherence of section-level summaries. Predictably, as the model's performance degrades, the reliability of high-abstraction-level chunks also diminishes. 

\section{Conclusion}
%In this paper, we introduce a novel Pre-RAG process step to enhance the retrieval process in LLMs by incorporating a Multiple Abstraction Level (MAL) RAG framework during document preprocessing. This approach enables a more comprehensive understanding by generating chunks at different abstraction levels, offering diverse and complementary perspectives when retrieving information. Additionally, we implement a similarity-based probability threshold using the Linq-Embed-Mistral embedding model~\cite{LinqAIResearch2024}, which effectively reduces noise and improves the relevance and accuracy of the generated answers.

%To support the evaluation of our approach, we have also released a dedicated Glyco-related dataset, which includes 800 question-answer pairs, enabling further benchmarking of retrieval-augmented generation systems. Our experimental results demonstrate that the MAL-RAG strategy significantly enhances the accuracy and relevance of the generated answers, contributing to the development of more robust and versatile retrieval-augmented generation systems.

In this work, we introduced a novel RAG approach that leverages MAL chunking to enhance information retrieval and Q/A performance. Our approach addresses the limitations of traditional single-level chunking methods by incorporating multiple levels of abstraction, ranging from multi-sentence-level to document-level, which allows LLMs to generate more accurate and coherent responses while mitigating the challenges associated with token limitations and the ‘lost in the middle’ problem. We demonstrated the effectiveness of our MAL-RAG framework in the under-explored scientific Glyco-domain, where it achieved a significant 25.739\% improvement in answer correctness compared to conventional single-level RAG methods. These results highlight the potential of our approach to enhance knowledge retrieval and adaptation in specialized domains where nuanced information processing is critical. We constructed a domain-specific Q/A dataset, which includes 800 curated Q/A pairs and can be used as a RAG-based Q/A benchmark. Future work will focus on further optimizing chunking strategies to better balance information density and relevance, exploring the applicability of our approach across broader scientific domains, and integrating advanced summarization techniques to further improve response accuracy and efficiency.

\bibliographystyle{named}
\bibliography{ijcai25}

\end{document}